\documentclass{article}

\PassOptionsToPackage{numbers, compress}{natbib}
\usepackage[final]{nips_2017}

\usepackage{float}
\usepackage[utf8]{inputenc} % allow utf-8 input
\usepackage[T1]{fontenc}    % use 8-bit T1 fonts
\usepackage{hyperref}       % hyperlinks
\usepackage{url}            % simple URL typesetting
\usepackage{booktabs}       % professional-quality tables
\usepackage{amsfonts}       % blackboard math symbols
\usepackage{nicefrac}       % compact symbols for 1/2, etc.
\usepackage{microtype}      % microtypography
\usepackage{enumitem}
\usepackage{graphicx}
\usepackage[british]{babel}
\usepackage{stackengine}
\usepackage{hhline}
\usepackage{multirow}
\usepackage{tabularx}
\usepackage{booktabs}
\usepackage{changepage}
\usepackage{amsmath}
\usepackage[thinlines]{easytable}
\restylefloat{table}
\setstackEOL{\#}

\title{ReabsNet: Detecting and Revising Adversarial Examples}
\author{
  Jiefeng Chen, Zihang Meng, Changtian Sun, Wei Tang, Yinglun Zhu \\
  University of Wisconsin - Madison\\
  \texttt{\{jiefeng, zihangm, changtiansun, weit, yinglun\}@cs.wisc.edu} 
}

\begin{document}
% \nipsfinalcopy is no longer used

\maketitle

\begin{abstract}
Though deep neural network has hit a huge success in recent studies and applications, it still remains vulnerable to adversarial perturbations which are imperceptible to humans. To address this problem, we propose a novel network called ReabsNet to achieve high classification accuracy in the face of various attacks. The approach is to augment an existing classification network with a guardian network to detect if a sample is natural or has been adversarially perturbed. Critically, instead of simply rejecting adversarial examples, we revise them to get their true labels. We exploit  the observation that a sample containing adversarial perturbations has a possibility of returning to its true class after revision. We demonstrate that our ReabsNet outperforms the state-of-the-art defense method under various adversarial attacks. 
\end{abstract}

\section{Introduction}

As a major technique used in machine learning, deep neural networks have shown good performance in many practical areas, especially in tasks involving pattern recognition and image classification \cite{nguyen2015deep}, such as image tagging \cite{li2016socializing}, face detecting and verification \cite{parkhi2015deep} and video classification \cite{yue2015beyond}. Although employing neural networks is one of the most accurate machine learning approaches, it can be quite vulnerable to adversarial examples whose aim is to attack the classifiers \cite{goodfellow2014explaining}. The \emph{adversarial perturbation} is defined as a test-time attack, which uses perturbed images that are visually the same as their unmodified versions, leading the neural network to generate a totally wrong labeling result. This problem has been shown to exist in almost all domains where the neural networks are used, strongly restricting the reliable applications of neural networks into many security-critical areas.

The research community has devoted much effort to building a neural network that is robust to adversarial examples. There are two major directions in the field of defending neural networks. One focuses on improving the model's robustness against adversarial perturbations itself. Another pays its attention to detecting adversarial examples and preventing them from attacking neural networks.

For the first direction, there exist theories and methods to improve the classification accuracy by strengthening the network structure and robustness (e.g., defensive distillation \cite{papernot2016distillation}, dropout \cite{feinman2017detecting}, convex outer adversarial polytope \cite{kolter2017provable}), training the network with enhanced training data \cite{miyato2015distributional}, etc. There is a possibility that these approaches would face difficulties, since systematic understandings of the network structures and adequate amount of adversarial training data are required \cite{lu2017safetynet}.

For the second direction, several different tracks for recent research works including training the neural network with an 'adversarial' class \cite{papernot2016limitations}, training an additional binary classifier \cite{gong2017adversarial, metzen2017detecting}, detecting adversarial examples based on Maximum Mean Discrepancy (MMD) \cite{papernot2016limitations}, observing the internal state of the classifier's late layers \cite{lu2017safetynet}, statistical analysis on Convolution Neural Network (CNN) layers \cite{li2016adversarial}, and so on. Such 'detecting and dropping' strategy has a limitation: if the detector misclassifies the natural images as adversarial ones, then these images are dropped. What's more, according to the definition that adversarial examples are samples that are closed to the natural ones but misclassified by the neural network, they should have the same labels with the corresponding natural ones. Thus simply rejecting them is not appropriate. 
%in a scenario that a neural network is an intermediate step of a whole process. For example, a following step might require the classified labels from neural network as input, while an adversary can feed the network by adversarial examples that get dropped by the neural net and thereby blocks the entire process.

To address these issues, we propose a reabsorption scheme and present a novel network called ReabsNet which can output a correct label for every input, even the adversarial ones. We build a \emph{guardian network} which can efficiently detect adversarial examples, and a modifier to modify detected adversarial examples iteratively until they are considered 'natural' by the guardian network. In the end, we enter all the natural and modified examples to the master network \cite{hinton2015distilling} for classification. The whole process resembles the physiology term \emph{reabsorption} \cite{wiki:reabsorption} that nutrient substances are reabsorbed from the tubule into the peritubular capillaries even if they are dropped due to various reasons at first.

% For evaluation, we test our method on MNIST dataset\cite{lecun1998mnist}. The experiment results show that, given an input image, a guardian network detects adversarial examples based on extracting and combining features of the last two ReLU layers. It can predict whether this input image is natural or adversarial with high confidence. Further, the images that are concluded as adversarial examples by the guardian network are tuned with an image modifier which works with similar mechanisms as attacking. The experiments show that selecting and constructing a master neural network is critical for the final classification accuracy, since the performance of a neural network classifier on all natural images might not be proportional to its performance on our amended natural images, i.e., we need to choose a network that matches our guardian network and image modifier and reaches the best performance when they are acting as a whole classification system. 

Our method focuses on developing an effective defense mechanism to adversarial examples, as well as building a strong classification system. Our main contributions are the following:

\begin{itemize}
\item We empirically prove the existence of a guardian network that can detect adversarial examples effectively.
\item We propose an image modification method based on attacking mechanisms, which is able to turn adversarial examples to natural ones so that a master network could correctly classify them. 
\item We perform an extensive experimental evaluation, showing that our method can successfully defend some existing attacks and outperform the state-of-the-art defense method. 
%We perform an extensive experimental evaluation, verify the performance and robustness of our model on different adversarial example generating methods. We come to the conclusion that: [to be continued]
% ToDo: finish the last line
\end{itemize}

\section{Background} \label{background}

Upon the demonstration of the idea \emph{adversarial examples} \cite{goodfellow2014explaining}, the study of these phenomena and the methods of dealing with this problem have been showing an uptrend. Given an input and its prospective label, after making perturbations to the input in a way where human eyes cannot tell the difference, it is possible that the classification result is different from the original label.

Currently, multiple kinds of attacks can achieve this goal to different extents, such as \emph{Fast Gradient Sign Method} \cite{goodfellow2014explaining}, \emph{DeepFool} \cite{moosavi2016deepfool}, \emph{$L_2$, $L_0$ and $L_{\infty}$ attack} \cite{carlini2016towards} and so on. And these methods will be introduced briefly in section~\ref{advAttacks}.

There are two general directions in tackling the problem. The first is to develop a network that is robust to one or several kinds of attacks. One example in this category is to use the convex outer adversarial polytope \cite{kolter2017provable}. This approach provides a method to train classifiers based on linear programming and duality theory. It uses a dual problem technique to represent the linear program as a deep network similar to back-propagation network. A decent proof is provided to show that this algorithm is robust to any norm-bounded adversarial attack. However, a small flaw in this method is that although it can be guaranteed to detect all the adversarial examples, some non-adversarial examples may also be misclassified. In addition, it is difficult to adjust this solution to a more complicated network due to high complexity of the algorithm.

In Madry et al.'s work \cite{madry2017towards}, they tried to propose a concrete \emph{guarantee} that a robust model should satisfy, and adapt their training methods towards it. The major steps involve specifying an attack model and raising a natural saddle point formulation. The resulting trained networks are shown to be robust against a wide range of attacks. They demonstrated that the saddle point problem can be solved eventually, but might not be in a reasonable time duration.

Another way of defense is to attach additional protection scheme to existing neural network models. Metzen et al. came up with the idea of adding an additional subnetwork specialized in detecting adversarial examples \cite{metzen2017detecting} to ResNet \cite{he2016deep}. Lu et al. designed a RBF-SVM based detector called SafetyNet with inputs of codes, of original/adversarial examples, in late ReLU layers of the main classification network, and this approach further enhances the detecting ability. \cite{lu2017safetynet} These are truly novel ideas in avoiding attacks by adversarial examples by using a detector. 

Apart from attaching networks for detecting adversarial examples, preprocessing the data samples are adopted in a manifold scheme \cite{wu2017manifold}. This scheme integrates protection into classification by using MCN or NCN model shells. Although further proof of the common manifold assumption is needed, this approach provides an advanced idea of preprocessing all the data points.

In our method, we adopted the second way of attaching additional protection scheme by using a guardian network. Different from the methods mentioned above, we make amendments for all the detected adversarial examples rather than simply dropping them.

\section{Method}
In this section, we briefly introduce the definition of adversarial examples and some adversarial attacks used in the experiment. Based on the defensive methods mentioned in Section \ref{background}, we further propose our novel scheme for correctly classifying adversarial perturbations.

\subsection{Problem Definition and Notation}
% definition of natural images and adversarial images.
% un/target attack
% black/white attack
Suppose we have a classifier $f_{\theta} (x)$ with parameters $\theta$ and outputs of discrete class labels. For a natural example $x$ and its ground truth label $y$, we want to train a proper $\theta$ that can make $f_{\theta}(x)=y$ stand with high probability. Further, it is reasonable to assume that $f_{\theta}(x+\epsilon)=f_{\theta}(x)$ when $\epsilon$ is small enough. However, many researchers \cite{szegedy2013intriguing, goodfellow2014explaining} have noticed small perturbations added to natural examples can confuse classifiers with high confidence and also in an incomprehensible way, specifically

\begin{equation}
f_{\theta}(x^{adv})=y^{adv} \neq y  \;\; \textrm{s.t.}\;\;  {||x^{adv}-x||}_p \leq \epsilon
\end{equation}

Here, $x^{adv}$ is an adversarial example to the classifier $f_{\theta}(x)$. If the class $y^{adv}$ is pre-specified by adversary, then we call $x^{adv}$ a \emph{targeted} adversarial example; otherwise, $x^{adv}$ can be viewed as an \emph{untargeted} adversarial example. 

\subsection{Adversarial Attacks} \label{advAttacks}

In this work, we consider three approaches for generating adversarial examples:

\textbf{Fast Gradient Sign Method \cite{goodfellow2014explaining}:~} This method is based on the assumption that neural networks are too linear to resist linear adversarial perturbation. Thus, the perturbation can be created using the sign of the elements of the gradient of the cost function with respect to the input. Adversarial examples can be given by:

\begin{equation}
x^{adv} = x + \epsilon \mbox{sign}(\nabla_x J(\theta,x,y))
\end{equation}

where $J(\theta,x,y)$ as the cost function of the classifier, $\theta$ is the parameters of the model, $x$ is the input, $y$ is the expected label and $\epsilon$ is the step size.

\textbf{CW $L_2$, $L_0$ and $L_{\infty}$ attack \cite{carlini2016towards}:~} It is a general way to quantify similarity with an $L_p$ norm form distance metric when defining adversarial examples. Given an image $x$, the $L_p$ attack aims to find a different image $x^{adv}$ that is similar to $x$ under $L_p$ distance, yet is given a totally different label by the classifier. In this paper, $L_2$ and $L_{\infty}$ attacks are used for experiment.

For $L_2$ attack, given a sample $x$ and a target class $t$, we search for $w$ that solves:

\begin{equation}
\begin{split}
\mbox{minimize~~} \|\frac{1}{2}(tanh(w)+1)-x\|_2^2+c\cdot f(\frac{1}{2}(tanh(w)+1)) \\
\mbox{with $f$ defined as~~} f(x')=\mbox{max}(\mbox{max}\{Z(x')_i:i\neq t\}-Z(x')_t,-\kappa)
\end{split}
\end{equation}

Here $f$ is based on the best objective function and the value of $\kappa$ is tried out for controlling the confidence.

For $L_\infty$ attack, an iterative attack is used and in each iteration, we need to solve

\begin{equation}
\mbox{minimize~~} c\cdot f(x+\delta)+\sum_{i}[(\delta_i-\tau)^+]
\end{equation}

Where $\delta_i$ is a penalization parameter in the $i$th iteration. After each iteration, if $\delta_i<\tau$, $\tau$ needs to be reduced to $0.9$ of its current value; otherwise, the search should be terminated.

% It is a general way to quantify similarity with an $L_p$ norm form distance metric when defining adversarial examples. These set of attacks can be used to construct an upper bound on the robustness of neural networks. Also, this work proved that the transferability, which defines the applicability of adversarial examples for one model on another one which are trained on different dataset, fails by constructing high-confidence adversarial examples on a unsecured model and then transfer to a secured model. Under $L_2$ distance, adversarial examples could be given by:
% \begin{equation}
% x_0^{adv} = x; \;\; x_{n+1}^{adv} = \textrm{Clip}_x^{\epsilon}\{ x_{n}^{adv} + \alpha sgn(\nabla_x J(x_n^{adv},f(x)))\}
% \end{equation}

\textbf{DeepFool  \cite{moosavi2016deepfool}:~} This is an untarget attacking technique optimized for the $L_p$ distance metric. When
implementing DeepFool, one can imagine that the neural networks are completely linear. Then the optimal solution towards this simplified problem can be derived iteratively and adversarial examples are thereby generated. This searching algorithm of DeepFool is greedy and it stops once the adversarial example crosses the classification boundary. Usually, DeepFool will generate adversarial examples closest to the original ones.

%the optimization strategy is strongly tied to existing optimization techniques. DeepFool has been proved to be efficient over various datasets and classifiers, and its ability to accurately evaluate the robustness of classifier also provides insight into the building of more robust classifier.

\subsection{Master Network} \label{master}
Master network is used to assign correct labels to input examples. The master network can just be a deep neural network. To get better performance in the master network, we also equip it with defensive distillation.

Here we briefly introduce the defensive distillation method \cite{papernot2016distillation}. Distillation \cite{hinton2015distilling} is a training procedure in which a softmax layer is added to the original DNN. This softmax layer considers the vector $Z(X)$ output by the last hidden layer and normalizes it into a probability vector $F(X)$ as the ultimate output of the whole network. The vector $F(X)$ assigns a probability to each class of the dataset for input $X$. Suppose there are $N$ classes in total, the output of neuron $i$ can be represented as the following:

\begin{equation}
\label{softmax}
F_i(X)=\left[ \frac{e^{z_i(X)/T}}{\Sigma_{l=0}^{N-1}{e^{z_l{(X)}/T}}} \right]_{i \in \{0 ... N-1\}}
\end{equation}

Here, $T$ is the \emph{temperature} parameter. The probability vectors produced by the new DNN are used as a \emph{soft labels}. The defensive distillation method just uses the same network architecture to train both the original network and the distilled network. 
%Denote $\mathcal{X}$ the training set. For a certain sample $X$, denote $Y(X)$ as its \emph{hard label} that represented by the form of one-hot encoding. For model $F$ with parameters $\theta_F$, the output is a conditional probability distribution $F(X)=p(\;\cdot\;|X,\theta_F)$. Now, for $X\in\mathcal{X}$, a new training set is constructed by samples in the form of $(X, F(X))$. Using the same neural network architecture as $F$, and the temperature still being $T$, a new \emph{distilled model} denoted as $F^d$ is trained. The advantage of using soft targets $F(X)$ is it makes use of additional knowledge about the relative differences between classes. This prevents models from fitting too tightly to the data.

\subsection{Guardian Network} \label{guardian}
We design a guardian network to detect adversarial examples from the natural ones. Before describing our guardian network elaborately, we first make the following assumption as the basis of the guardian network.

\textbf{Assumption:~} Distributions of natural examples and adversarial ones are statistically different, and this difference can be learned by the guardian network.

We design a two-category deep neural network, with both natural and adversarial examples as input, to be our guardian network. Suppose $x$ is an input example, we will get an output $G(x) \in [0,1]^2$ in the softmax layer showing the probability of classifying $x$ as a natural example and an adversarial example. If the probability of an input being an adversarial example is higher, our guardian network will put it into a modifier that can turn it back to a 'natural' example. Otherwise, our guardian network will send it into the master network directly. Since it is very difficult for adversarial examples to be both misclassified by the master network and the guardian network, they together form a very reliable classifier which can classify not only natural examples, but also adversarial examples.

To train this guardian network, we first need to train our master network with natural examples. Subsequently, we generate adversarial data for each natural examples in the training dataset through one of the attacking methods discussed in Section~\ref{advAttacks}. Finally, we train our guardian network with a balanced binary classification dataset of twice the size of the original one.

\begin{figure} 
	\includegraphics[width=\linewidth]{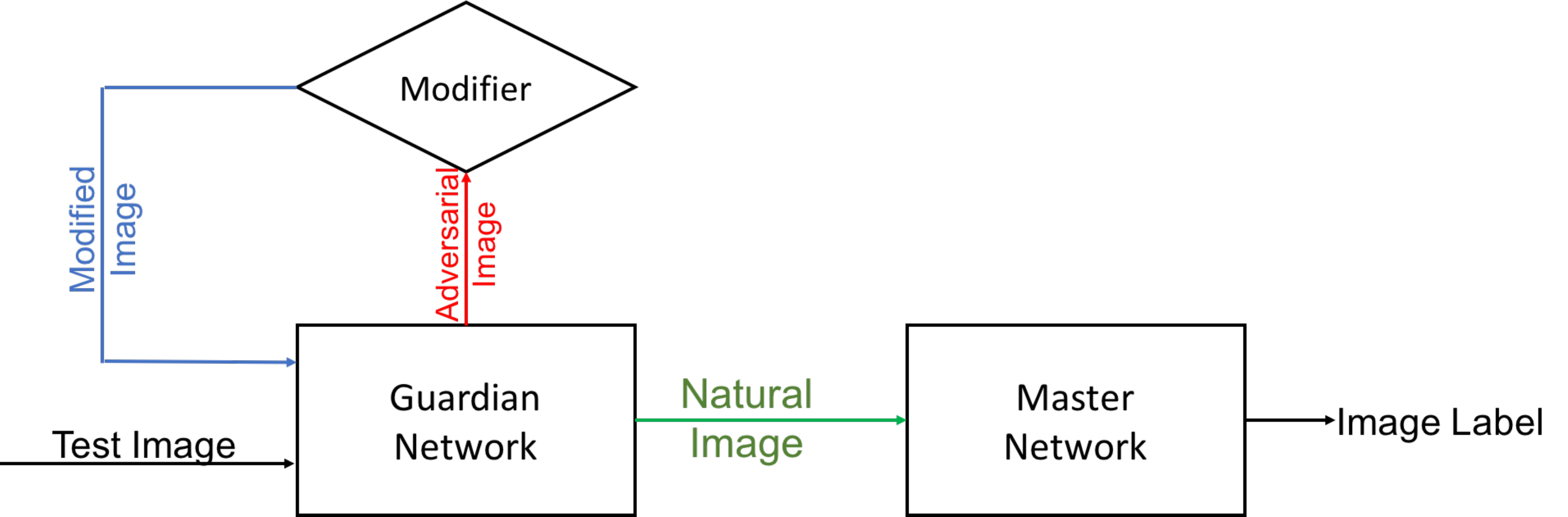}
    \caption{Structure of ReabsNet: there are three parts in our ReabsNet, the guardian network, the modifier and the master network. When a test image is received by ReabsNet, it is first examined by the guardian network: if it is classified as an adversarial image, it will be sent to the modifier; if it is classified as natural image, it will be sent to the master network for classification. The job of modifier is, for each adversarial image it received from guardian network, it amends the image iteratively until the modified image can pass the guardian network, i.e., be classified as natural image by the guardian network.}
    \label{fig:flowchart1}
\end{figure}

\subsection{Revising Adversarial Examples}
% more assumptions here, like the distribution between real exs and the fake ones.

Our guardian network mentioned in Section~\ref{guardian} can detect many possible adversarial examples. In previous approaches (e.g., the SafetyNet \cite{lu2017safetynet}), people just prevent these adversarial examples from entering the main classification network. For improvement, we move forward by modifying the adversarial examples and then enter the \emph{revised adversarial examples} into the master network. 

With the assumption in Section~\ref{guardian}, a good guardian network can be obtained. If the guardian network reckons an input as a natural example, the master network should be able to classify it correctly. Thus, we can slightly modify the adversarial example so that the guardian network may classify it as a natural example. Afterwards, it will be entered into the master network. This modification method can be the one of the attacks described in Section~\ref{advAttacks}. We show below why this method can work. 

Suppose $x$ is the original natural example, and $x^{adv}$ is the corresponding adversarial example detected by the guardian network. With the property of attack under $p$ norm, we have

\begin{equation}
{||x-x^{adv}||}_p \leq \epsilon_1
\label{attack}
\end{equation}

We further suppose $x^{ame}$ is the amendment of the adversarial example $x^{adv}$, which means that $x^{ame}$ is viewed as natural example by the guardian network. Based on the assumption, it is likely that the master network can classify $x^{ame}$ correctly. Besides, we have

\begin{equation}
{||x^{ame}-x^{adv}||}_p \leq \epsilon_2
\label{amend}
\end{equation}

Combining equation~\eqref{attack} and equation~\eqref{amend}, we have

\begin{equation}
{||x^{ame}-x||}_p \leq {||x^{ame}-x^{adv}||}_p+{||x-x^{adv}||}_p \leq \epsilon_2 + \epsilon_1
\label{tworeal}
\end{equation}

% ToDo: explanation of the letter 'p'
% explained above as p norm

In equation~\eqref{tworeal}, although $\epsilon_1$ is determined by the attackers, we can choose our modification strategy to control $\epsilon_2$. In order to control the distance between $x^{ame}$ and $x$ in worst case, DeepFool \cite{moosavi2016deepfool} can be used here. In this way, after amending the adversarial example, we obtain a natural example lying very close to the original one. Thus the label of $x^{ame}$ should be the same as $x$. Hence, through revising the adversarial with respect to the guardian network, we find a way to correctly classify adversarial examples, and this can be viewed as a novel defending mechanism.

\subsection{Reabsorption Network}

% ToDo: flow chart. can also be added in the front of this paper. Modify the next figure 1
We name our entire model as \emph{Reabsorption Network} (shortened as \emph{ReabsNet}). Because our classification process resembles the renal physiological process \emph{Reabsorption} in which nutrient substances like glucose and amino acids are reabsorbed after they are dropped at first. In our method, we actually revise adversarial examples into natural ones and then re-classify them rather than simply 'drop them out of the body'.

The structure of our model is demonstrated in figure~\ref{fig:flowchart1}. The model consists of a guardian network, an image modifier and a master network that is used for classification. If the guardian network concludes an image as an adversarial image, this sample is then modified to become 'natural'. The image can be sent to the master network for classification only when it seems natural to the guardian network.

\section{Experiment}
%statistics needed:
%1. classification network accuracy;
%2. attack accuracy w.r.t. original network; distilled one; and distilled one plus detector
%3. classification accuracy of amendment examples
%In this section, we evaluate our method and present the results. 
\subsection{Dataset}
The dataset that we use is MNIST. MNIST \cite{lecun1998mnist} is a popular benchmark dataset for computer vision tasks. It consists of a training set of $60,000$ examples, and a testing set of $10,000$ examples, where each $28 \times 28$ image belongs to one of ten handwritten digits (from $0$ to $9$). In our experiments, we use $55,000$ examples in the training set to train models and use the remaining $5,000$ examples to validate models. We also normalize each pixel's value to the range of $[-0.5,0.5]$ before feeding the image into the network. 

\subsection{Implementing Details}
\begin{figure}
	\includegraphics[width=\linewidth]{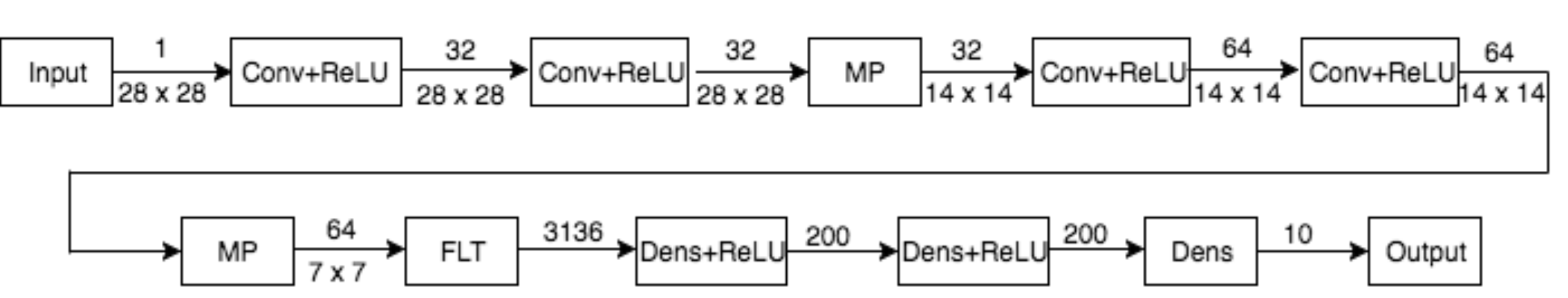}
    \caption{Structure of Master Network. In this graph, 'Conv+ReLU' stands for a convolutional layer with a ReLU activation layer. 'MP' stands a for Max-Pooling layer. 'Dens' denotes a fully-connected layer, and 'Dens+ReLU' stands for a fully-connected layer with a ReLU activation layer. 'FLT' denotes a flattening operation. The numbers on top of arrows denote the numbers of feature maps, and those below arrows denote spatial resolution.}
    \label{fig:MasterStructure}
\end{figure}

\begin{figure}
\label{detector}
	\includegraphics[width=\linewidth]{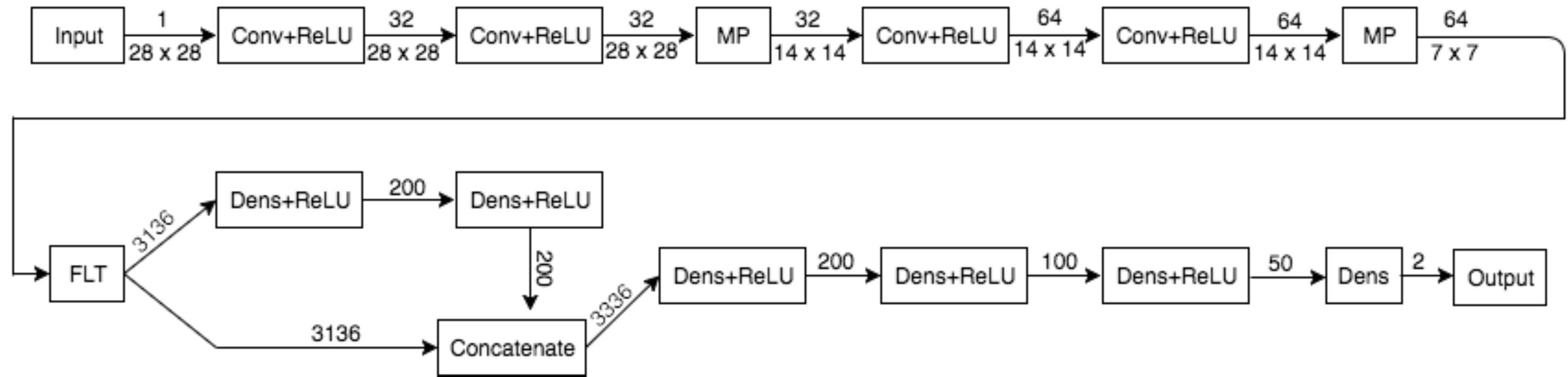}
    \caption{Structure of Guardian Network. 'Concatenate' is a step that concatenates features from two steps together. The other symbols and notations have the same representations as figure~\ref{fig:MasterStructure}}
    \label{fig:DetectorStructure}
\end{figure}

The structure of the master network is showed in figure~\ref{fig:MasterStructure}. For the defensive distillation training, we set $T=100$. The structure of the guardian network is showed in figure~\ref{fig:DetectorStructure}. We use DeepFool method to attack the master network and generate adversarial examples. Then we use them to train the guardian network. Afterwards, DeepFool method is used again to attack the guardian network so as to turn the adversarial examples back to the natural images. In consideration of the computing time, we generate $1000$ possible adversarial examples with FGSM, DeepFool, CW $L_2$ attack (targetedly/untargetedly) respectively, and generate $100$ possible adversarial examples using CW $L_{\infty}$ attack (targetedly/untargetedly) to calculate the defense success rate under these attacks. 

\subsection{Results}
To evaluate the performance of our network against adversarial perturbations, we conduct experiments on MNIST and report results under several common effective adversarial attack methods.

\begin{figure}
\centering
  \begin{minipage}[b]{0.15\linewidth}
  \centering
  \centerline{}
  \end{minipage}
  \hspace{20pt}
  \vspace{1.5pt}
  \begin{minipage}[b]{0.15\linewidth}
  \centering
  \centerline{Natural}
  \end{minipage}
  \hspace{20pt}
  \vspace{1.5pt}
  \begin{minipage}[b]{0.15\linewidth}
  \centering
  \centerline{Adversarial}
  \end{minipage}
  \hspace{20pt}
  \vspace{1.5pt}
  \begin{minipage}[b]{0.15\linewidth}
  \centering
  \centerline{Modified}
  \end{minipage}
  \hspace{20pt}
  \vspace{1.5pt}
  \vfill
  \begin{minipage}[b]{0.15\linewidth}
  \centering
  \centerline{FGSM}
  \centerline{}
  \centerline{}
  \centerline{}
  \centerline{}
  \end{minipage}
  \hspace{20pt}
  \vspace{1.5pt}
  \begin{minipage}[b]{0.15\linewidth}
  \centering
  \centerline{\includegraphics[width=1.25\textwidth]{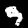}}
  \centerline{\footnotesize{$prediction = 9$}}
  \end{minipage}
  \hspace{20pt}
  \vspace{1.5pt}
  \begin{minipage}[b]{0.15\linewidth}
  \centering
  \centerline{\includegraphics[width=1.25\textwidth]{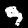}}
  \centerline{\footnotesize{$prediction = 3$}}
  \end{minipage}
  \hspace{20pt}
  \vspace{1.5pt}
  \begin{minipage}[b]{0.15\linewidth}
  \centering
  \centerline{\includegraphics[width=1.25\textwidth]{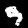}}
  \centerline{\footnotesize{$prediction = 9$}}
  \end{minipage}
  \hspace{20pt}
  \vspace{1.5pt}
  \vfill
  \begin{minipage}[b]{0.15\linewidth}
  \centering
  \centerline{DeepFool}
  \centerline{}
  \centerline{}
  \centerline{}
  \centerline{}
  \end{minipage}
  \hspace{20pt}
  \vspace{1.5pt}
  \begin{minipage}[b]{0.15\linewidth}
  \centering
  \centerline{\includegraphics[width=1.25\textwidth]{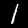}}
  \centerline{\footnotesize{$prediction = 1$}}
  \end{minipage}
  \hspace{20pt}
  \vspace{1.5pt}
  \begin{minipage}[b]{0.15\linewidth}
  \centering
  \centerline{\includegraphics[width=1.25\textwidth]{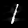}}
  \centerline{\footnotesize{$prediction = 8$}}
  \end{minipage}
  \hspace{20pt}
  \vspace{1.5pt}
  \begin{minipage}[b]{0.15\linewidth}
  \centering
  \centerline{\includegraphics[width=1.25\textwidth]{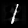}}
  \centerline{\footnotesize{$prediction = 1$}}
  \end{minipage}
  \hspace{20pt}
  \vspace{1.5pt}
  \vfill
  \begin{minipage}[b]{0.15\linewidth}
  \centering
  \centerline{CW $L_2$}
  \centerline{}
  \centerline{}
  \centerline{}
  \centerline{}
  \end{minipage}
  \hspace{20pt}
  \vspace{1.5pt}
  \begin{minipage}[b]{0.15\linewidth}
  \centering
  \centerline{\includegraphics[width=1.25\textwidth]{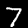}}
  \centerline{\footnotesize{$prediction = 7$}}
  \end{minipage}
  \hspace{20pt}
  \vspace{1.5pt}
  \begin{minipage}[b]{0.15\linewidth}
  \centering
  \centerline{\includegraphics[width=1.25\textwidth]{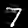}}
  \centerline{\footnotesize{$prediction = 9$}}
  \end{minipage}
  \hspace{20pt}
  \vspace{1.5pt}
  \begin{minipage}[b]{0.15\linewidth}
  \centering
  \centerline{\includegraphics[width=1.25\textwidth]{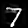}}
  \centerline{\footnotesize{$prediction = 7$}}
  \end{minipage}
  \hspace{20pt}
  \vspace{1.5pt}
  \vfill
  \begin{minipage}[b]{0.15\linewidth}
  \centering
  \centerline{CW $L_\infty$}
  \centerline{}
  \centerline{}
  \centerline{}
  \centerline{}
  \end{minipage}
  \hspace{20pt}
  \vspace{1.5pt}
  \begin{minipage}[b]{0.15\linewidth}
  \centering
  \centerline{\includegraphics[width=1.25\textwidth]{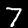}}
  \centerline{\footnotesize{$prediction = 7$}}
  \end{minipage}
  \hspace{20pt}
  \vspace{1.5pt}
  \begin{minipage}[b]{0.15\linewidth}
  \centering
  \centerline{\includegraphics[width=1.25\textwidth]{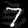}}
  \centerline{\footnotesize{$prediction = 9$}}
  \end{minipage}
  \hspace{20pt}
  \vspace{1.5pt}
  \begin{minipage}[b]{0.15\linewidth}
  \centering
  \centerline{\includegraphics[width=1.25\textwidth]{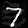}}
  \centerline{\footnotesize{$prediction = 7$}}
  \end{minipage}
  \hspace{20pt}
  \vspace{1.5pt}
\caption{Examples of natural, adversarial and modified-adversarial MNIST images. Adversarial images are generated by FGSM, DeepFool, $CW\ L_2$ and $CW\ L_{\infty}$ respectively. The prediction of each image by our ReabsNet is marked at the bottom of the image. After our image modification step, the modified-adversarial examples can be predicted correctly again.}
\label{fig:mnistSamples}
\end{figure}

\subsubsection{Performance of Guardian Network}
Before discussing the whole network’s performance, we first report the performance of our guardian network. Our guardian network can detect adversarial examples with high success rate on MNIST. Table~\ref{tab:guardianTable} shows the guardian network’s performance on various attack methods and under a non-attack situation. 

The experimental results show that our guardian network is able to learn the boundary between natural examples and adversarial ones and testify that the distributions of these two are different. In addition, the efficiency of our guardian network is critical to the further modification and classification steps.

\begin{table}[b]
	
	\begin{adjustwidth}{-0.5cm}{}
	\begin{tabular}{m{2cm} | m{1.4cm}  m{1.2cm}  m{1.4cm}  m{1.2cm}  m{1.4cm} m{1.2cm}}
		\toprule
		\multicolumn{1}{c|}{Attack Method} & \multicolumn{1}{c}{Non-Attack} & \multicolumn{1}{c}{DeepFool} & \multicolumn{1}{c}{$L_2$ Untargeted} & \multicolumn{1}{c}{$L_2$ Targeted} & \multicolumn{1}{c}{$L_{\infty}$ Untargeted} & \multicolumn{1}{c}{$L_{\infty}$ Targeted} \\
		\midrule
		\multicolumn{1}{c|}{Detect Success Rate} & \multicolumn{1}{c}{0.9926}  & \multicolumn{1}{c}{0.980} & \multicolumn{1}{c}{0.983} & \multicolumn{1}{c}{0.999} & \multicolumn{1}{c}{0.99} & \multicolumn{1}{c}{1.0} \\	
		\bottomrule
	\end{tabular}
	\end{adjustwidth}
	\caption{Guardian net's success rate on detecting adversarial images generated from different attack methods. Note: we don't include the result of FGSM attack because, we only have 7 adversarial test images generated from FGSM, which means the resulted success rate might not be accurate.}
    \label{tab:guardianTable}
\end{table}
%We also show that $L_{\infty}$ attack is easier to be detected than $L_2$ and DeepFool attacks, as a result of more obvious perturbation patterns of $L_{\infty}$ attack.

\begin{table}[!htb]
	\centering
	\begin{tabular}{m{2.6cm} | m{3cm} |  m{2cm}}
		\toprule
		\multirow{2}{3cm}{\textbf{  Attack Method}} & \multicolumn{2}{c}{\textbf{Defense Success Rate}}\\
		\cmidrule{2-3}
		{} & \multicolumn{1}{c|}{\textbf{Master Network Only}} & \multicolumn{1}{c}{\textbf{ReabsNet}}\\
		\midrule
		%\multicolumn{1}{c|}{\textbf{Non-Attack}} & \multicolumn{1}{c|}{0.9905} & \multicolumn{1}{c}{0.9905}\\
        
		\multicolumn{1}{c|}{\textbf{FGSM}} & \multicolumn{1}{c|}{0.998} & \multicolumn{1}{c}{1.0}\\
		
		\multicolumn{1}{c|}{\textbf{DeepFool}} & \multicolumn{1}{c|}{0.0} & \multicolumn{1}{c}{0.709}\\
		
		\multicolumn{1}{c|}{\textbf{CW $L_2$ Untargeted}} & \multicolumn{1}{c|}{0.001} & \multicolumn{1}{c}{0.983}\\

		\multicolumn{1}{c|}{\textbf{CW $L_2$ Targeted}} & \multicolumn{1}{c|}{0.001} & \multicolumn{1}{c}{0.962}\\

		\multicolumn{1}{c|}{\textbf{CW $L_{\infty}$ Untargeted}} & \multicolumn{1}{c|}{0.0} & \multicolumn{1}{c}{0.99}\\

		\multicolumn{1}{c|}{\textbf{CW $L_{\infty}$ Targeted}} & \multicolumn{1}{c|}{0.0} & \multicolumn{1}{c}{0.95}\\

		\bottomrule
	\end{tabular}
	\caption{Defense success rate on adversarial images generated from different methods. We don't set a distance bound here because the attack algorithms will try to find the minimum distance between the natural image and the adversarial one, and the distance metrics that different attack methods use may vary. The defense success rate is evaluated on six different attack methods: FGSM, DeepFool, CW $L_2$ Untargeted, CW $L_2$ Targeted, CW $L_{\infty}$ Untargeted and CW $L_{\infty}$ Targeted. And the table shows the resulted defense success rates under the cases of using the master network only and using the whole ReabsNet, respectively.}
\label{tab:table2}
\end{table}

\subsubsection{Performance of ReabsNet}
Our network achieves high classification accuracy under several common effective attack methods on MNIST. Both our master network and ReabsNet can achieve 99.05\% classification accuracy on natural images. Table~\ref{tab:table2} shows that our ReabsNet still achieves high classification accuracy under various attacks and is prominently better than the model (our master network) with only defensive distillation training. Some examples of natural, adversarial and modified adversarial images can be seen in figure \ref{fig:mnistSamples}.

The results testify that an adversarial example which is misclassified by the master network has the possibility of returning to the true class after certain modification. Since we use DeepFool to generate adversarial examples for training the detector and also as the modification method under all the different attack methods, it demonstrates the generalization ability of our network across different attacks. In this way, at training time, we don’t have to know what attacks will occur at test time, which is exactly the case in reality.

\subsubsection{Comparing with Madry's Model}
\begin{table}[b]\small
	\begin{adjustwidth}{-1.8cm}{}
	\begin{tabular}{m{1.7cm} | m{0.8cm} m{0.6cm} m{0.8cm} m{0.8cm} m{0.9cm} m{0.9cm}| m{0.8cm} m{0.6cm} m{0.8cm} m{0.8cm} m{0.9cm} m{0.9cm}}
		\toprule
        {} & \multicolumn{12}{c}{\textbf{Defense Success Rate}}\\
        \cmidrule{2-13}
		{}  & \multicolumn{6}{c|}{\textbf{$\epsilon < 0.3$}} & \multicolumn{6}{c}{\textbf{$\epsilon < 0.6$}}\\
		\cmidrule{2-13} 
		{} & \textbf{FGSM} & \textbf{ DF} & \textbf{$L_2$ U} & \textbf{$L_2$ T} & \textbf{$L_{\infty}$ U} & \textbf{$L_{\infty}$ T} & \textbf{FGSM} & \textbf{ DF} & \textbf{$L_2$ U} & \textbf{$L_2$ T} & \textbf{$L_{\infty}$ U} & \textbf{$L_{\infty}$ T} \\
		\midrule
		\textbf{Our Model} & \multicolumn{1}{c}{1.0} & \text{0.92} & \text{0.984} & \text{0.996} & \text{0.99} & \multicolumn{1}{c|}{\text{0.95}} & \multicolumn{1}{c}{1.0} & \text{0.744} & \text{0.983} & \text{0.973} & \text{0.99} & \text{0.95}\\
		\multicolumn{1}{c|}{\textbf{Madry's Model}} & \multicolumn{1}{c}{1.0} & \text{0.968} & \text{0.962} & \text{0.997} & \text{0.93} & \multicolumn{1}{c|}{\text{0.98}} & \multicolumn{1}{c}{1.0} & \text{0.946} & \text{0.582} & \text{0.226} & \text{0.15} & \text{0.43}\\
		\bottomrule
	\end{tabular}
	\end{adjustwidth}
	\caption{Compare Defense Success Rate with Madry's Model, using different attack methods and under perturbation scales $\epsilon<0.3$ and $\epsilon<0.6$. In this table, DF stands for the DeepFool attack, $L_2\ U$ stands for the CW $L_2$ norm untarget attack, $L_2\ T$ stands for the CW $L_2$ norm Targeted attack, $L_{\infty}\ U$ stands for the CW $L_{\infty}$ norm untarget attack, and $L_{\infty}\ T$ stands for the CW $L_{\infty}$ norm Targeted attack.}
    \label{compare}
\end{table}

We further use a state-of-the-art defense method described in Madry's paper \cite{madry2017towards} as our baseline to evaluate the performance of ReabsNet.

Madry et al. trained a robust network on the MNIST dataset based on the method described in their paper \cite{madry2017towards}. They also posted a MNIST Adversarial Examples Challenge \cite{madry2017challenge} to allow others to attack their model. From the leaderboards, we can see their model is very robust to adversarial attacks. However, their network is trained against an iterative adversary that is allowed to perturb each pixel value (in the range of $[-0.5,0.5]$) by at most $\epsilon=0.3$. In consideration of fairness, when comparing the results, we also set a $L_{\infty}$ distance bound in the attack algorithms: if the $L_{\infty}$ distance between the original image and the adversarial image found is larger than $\epsilon$, we say the attack algorithms fail to find a valid adversarial example and replace the adversarial image with the original one.

The results of experiments on our ReabsNet and Madry's model under various attacks appear in Table~\ref{compare}. Our model is comparable with Madry's model under the restricted condition they specify, and tends to be consistently better when the perturbation scale is relaxed beyond 0.3, where the classification task becomes harder. 

\section{Discussion}
In this paper, we have described a novel network, ReabsNet, with high classification ability that can correctly classify both natural and adversarial examples. It can detect and classify the adversarial examples from attacking methods not seen in the previous training process. Despite good performance on the attacks to master networks, we have not addressed the situation where the attacker also knows the architecture and parameters of the guardian network and then attack the master network and guardian network at the same time. We leave this to future work. Finally, we believe that our technique of leveraging the guardian network and modifier could be applied to help understand the distributions of natural and adversarial examples, which could be an interesting direction for the future work.

\subsubsection*{Acknowledgments}
We thank Prof. Yingyu Liang for his insightful instruction and providing us with such a great opportunity to explore more in the field of machine learning.
\bibliography{nips_2017}

% Generated by IEEEtran.bst, version: 1.14 (2015/08/26)
\begin{thebibliography}{10}
\providecommand{\url}[1]{#1}
\csname url@samestyle\endcsname
\providecommand{\newblock}{\relax}
\providecommand{\bibinfo}[2]{#2}
\providecommand{\BIBentrySTDinterwordspacing}{\spaceskip=0pt\relax}
\providecommand{\BIBentryALTinterwordstretchfactor}{4}
\providecommand{\BIBentryALTinterwordspacing}{\spaceskip=\fontdimen2\font plus
\BIBentryALTinterwordstretchfactor\fontdimen3\font minus
  \fontdimen4\font\relax}
\providecommand{\BIBforeignlanguage}[2]{{%
\expandafter\ifx\csname l@#1\endcsname\relax
\typeout{** WARNING: IEEEtran.bst: No hyphenation pattern has been}%
\typeout{** loaded for the language `#1'. Using the pattern for}%
\typeout{** the default language instead.}%
\else
\language=\csname l@#1\endcsname
\fi
#2}}
\providecommand{\BIBdecl}{\relax}
\BIBdecl

\bibitem{nguyen2015deep}
A.~Nguyen, J.~Yosinski, and J.~Clune, ``Deep neural networks are easily fooled:
  High confidence predictions for unrecognizable images,'' in \emph{Proceedings
  of the IEEE Conference on Computer Vision and Pattern Recognition}, 2015, pp.
  427--436.

\bibitem{li2016socializing}
X.~Li, T.~Uricchio, L.~Ballan, M.~Bertini, C.~G. Snoek, and A.~D. Bimbo,
  ``Socializing the semantic gap: A comparative survey on image tag assignment,
  refinement, and retrieval,'' \emph{ACM Computing Surveys (CSUR)}, vol.~49,
  no.~1, p.~14, 2016.

\bibitem{parkhi2015deep}
O.~M. Parkhi, A.~Vedaldi, A.~Zisserman \emph{et~al.}, ``Deep face
  recognition.'' in \emph{BMVC}, vol.~1, no.~3, 2015, p.~6.

\bibitem{yue2015beyond}
J.~Yue-Hei~Ng, M.~Hausknecht, S.~Vijayanarasimhan, O.~Vinyals, R.~Monga, and
  G.~Toderici, ``Beyond short snippets: Deep networks for video
  classification,'' in \emph{Proceedings of the IEEE conference on computer
  vision and pattern recognition}, 2015, pp. 4694--4702.

\bibitem{goodfellow2014explaining}
I.~J. Goodfellow, J.~Shlens, and C.~Szegedy, ``Explaining and harnessing
  adversarial examples,'' \emph{arXiv preprint arXiv:1412.6572}, 2014.

\bibitem{papernot2016distillation}
N.~Papernot, P.~McDaniel, X.~Wu, S.~Jha, and A.~Swami, ``Distillation as a
  defense to adversarial perturbations against deep neural networks,'' in
  \emph{Security and Privacy (SP), 2016 IEEE Symposium on}.\hskip 1em plus
  0.5em minus 0.4em\relax IEEE, 2016, pp. 582--597.

\bibitem{feinman2017detecting}
R.~Feinman, R.~R. Curtin, S.~Shintre, and A.~B. Gardner, ``Detecting
  adversarial samples from artifacts,'' \emph{arXiv preprint arXiv:1703.00410},
  2017.

\bibitem{kolter2017provable}
J.~Z. Kolter and E.~Wong, ``Provable defenses against adversarial examples via
  the convex outer adversarial polytope,'' \emph{arXiv preprint
  arXiv:1711.00851}, 2017.

\bibitem{miyato2015distributional}
T.~Miyato, S.-i. Maeda, M.~Koyama, K.~Nakae, and S.~Ishii, ``Distributional
  smoothing with virtual adversarial training,'' \emph{arXiv preprint
  arXiv:1507.00677}, 2015.

\bibitem{lu2017safetynet}
J.~Lu, T.~Issaranon, and D.~Forsyth, ``Safetynet: Detecting and rejecting
  adversarial examples robustly,'' \emph{arXiv preprint arXiv:1704.00103},
  2017.

\bibitem{papernot2016limitations}
N.~Papernot, P.~McDaniel, S.~Jha, M.~Fredrikson, Z.~B. Celik, and A.~Swami,
  ``The limitations of deep learning in adversarial settings,'' in
  \emph{Security and Privacy (EuroS\&P), 2016 IEEE European Symposium
  on}.\hskip 1em plus 0.5em minus 0.4em\relax IEEE, 2016, pp. 372--387.

\bibitem{gong2017adversarial}
Z.~Gong, W.~Wang, and W.-S. Ku, ``Adversarial and clean data are not twins,''
  \emph{arXiv preprint arXiv:1704.04960}, 2017.

\bibitem{metzen2017detecting}
J.~H. Metzen, T.~Genewein, V.~Fischer, and B.~Bischoff, ``On detecting
  adversarial perturbations,'' \emph{arXiv preprint arXiv:1702.04267}, 2017.

\bibitem{li2016adversarial}
X.~Li and F.~Li, ``Adversarial examples detection in deep networks with
  convolutional filter statistics,'' \emph{arXiv preprint arXiv:1612.07767},
  2016.

\bibitem{hinton2015distilling}
G.~Hinton, O.~Vinyals, and J.~Dean, ``Distilling the knowledge in a neural
  network,'' \emph{arXiv preprint arXiv:1503.02531}, 2015.

\bibitem{wiki:reabsorption}
\BIBentryALTinterwordspacing
``Reabsorption.'' [Online]. Available:
  \url{https://en.wikipedia.org/wiki/Reabsorption}
\BIBentrySTDinterwordspacing

\bibitem{moosavi2016deepfool}
S.-M. Moosavi-Dezfooli, A.~Fawzi, and P.~Frossard, ``Deepfool: a simple and
  accurate method to fool deep neural networks,'' in \emph{Proceedings of the
  IEEE Conference on Computer Vision and Pattern Recognition}, 2016, pp.
  2574--2582.

\bibitem{carlini2016towards}
N.~Carlini and D.~Wagner, ``Towards evaluating the robustness of neural
  networks,'' \emph{arXiv preprint arXiv:1608.04644}, 2016.

\bibitem{madry2017towards}
A.~Madry, A.~Makelov, L.~Schmidt, D.~Tsipras, and A.~Vladu, ``Towards deep
  learning models resistant to adversarial attacks,'' \emph{arXiv preprint
  arXiv:1706.06083}, 2017.

\bibitem{he2016deep}
K.~He, X.~Zhang, S.~Ren, and J.~Sun, ``Deep residual learning for image
  recognition,'' in \emph{Proceedings of the IEEE conference on computer vision
  and pattern recognition}, 2016, pp. 770--778.

\bibitem{wu2017manifold}
X.~Wu, U.~Jang, L.~Chen, and S.~Jha, ``Manifold assumption and defenses against
  adversarial perturbations,'' \emph{arXiv preprint arXiv:1711.08001}, 2017.

\bibitem{szegedy2013intriguing}
C.~Szegedy, W.~Zaremba, I.~Sutskever, J.~Bruna, D.~Erhan, I.~Goodfellow, and
  R.~Fergus, ``Intriguing properties of neural networks,'' \emph{arXiv preprint
  arXiv:1312.6199}, 2013.

\bibitem{lecun1998mnist}
Y.~LeCun, ``The mnist database of handwritten digits,'' \emph{http://yann.
  lecun. com/exdb/mnist/}, 1998.

\bibitem{madry2017challenge}
{Madry, Aleksander and Makelov, Aleksandar and Schmidt, Ludwig and Tsipras,
  Dimitris and Vladu, Adrian}, ``{MNIST C}hallenge,''
  \url{https://github.com/MadryLab/mnist_challenge}.

\end{thebibliography}

\end{document}